\documentclass[runningheads]{llncs}
\usepackage{graphicx}
\usepackage{amsmath,amssymb} 
\usepackage{color}

\usepackage{epsfig}
\usepackage{graphicx}
\usepackage{subfigure}
\usepackage{microtype}
\usepackage{multirow}
\usepackage{enumitem}
\usepackage{eso-pic}
\usepackage{bm}
\usepackage{booktabs}
\usepackage{adjustbox}
\usepackage{hyperref}

\begin{document}

\title{Quantized Densely Connected U-Nets for Efficient Landmark Localization}

\titlerunning{Quantized Densely Connected U-Nets for Efficient Landmark Localization}
%
\author{Zhiqiang Tang$^1$ \and
Xi Peng$^2$ \and
Shijie Geng$^1$ \and Lingfei Wu$^3$ \and \\Shaoting Zhang$^4$ \and Dimitris Metaxas$^1$}
%
\authorrunning{Tang et al.}
%

\institute{
$^1$Rutgers University, 
\email{\{zt53, sg1309, dnm\}@rutgers.edu}\\
$^2$Binghamton University, \email{xpeng@binghamton.edu}\\
$^3$IBM T. J. Watson, 
\email{lwu@email.wm.edu}\\
$^4$SenseTime,
\email{zhangshaoting@sensetime.com}\\
}

\maketitle

\begin{abstract}

  In this paper, we propose quantized densely connected U-Nets for efficient visual landmark localization. The idea is that features of the same semantic meanings are globally reused across the stacked U-Nets. This dense connectivity largely improves the information flow, yielding improved localization accuracy. However, a vanilla dense design would suffer from critical efficiency issue in both training and testing. To solve this problem, we first propose order-K dense connectivity to trim off long-distance shortcuts; then, we use a memory-efficient implementation to significantly boost the training efficiency and investigate an iterative refinement that may slice the model size in half. 
  Finally, to reduce the memory consumption and high precision operations both in training and testing, we further quantize weights, inputs, and gradients of our localization network to low bit-width numbers.
  We validate our approach in two tasks: human pose estimation and face alignment. The results show that our approach achieves state-of-the-art localization accuracy, but using $\sim$70\% fewer parameters, $\sim$98\% less model size and saving $\sim$75\% training memory compared with other benchmark localizers. The code is available at \href{https://github.com/zhiqiangdon/CU-Net}{https://github.com/zhiqiangdon/CU-Net}. 
  
  
\end{abstract}
\section{Introduction}

Locating visual landmarks, such as human body joints \cite{toshev2014deeppose} and facial key points \cite{xiong2013supervised}, is an important yet challenging problem. The stacked U-Nets, {\it e.g.} hourglasses (HGs) \cite{newell2016stacked}, are widely used in landmark localization. Generally speaking, their success can be attributed to design patterns: 1) within each U-Net, connect the top-down and bottom-up feature blocks to encourage gradient flow; and 2) stack multiple U-Nets in a cascade to refine prediction stage by stage.

However, the shortcut connection exists only ``locally'' inside each U-Net \cite{ronneberger2015u}. There is no ``global'' connection across U-Nets except the cascade. Blocks in different U-Nets cannot share features, which may impede the information flow and lead to redundant parameters.

\begin{figure*}[t!]
\centering
  \includegraphics[width=1.0\linewidth]{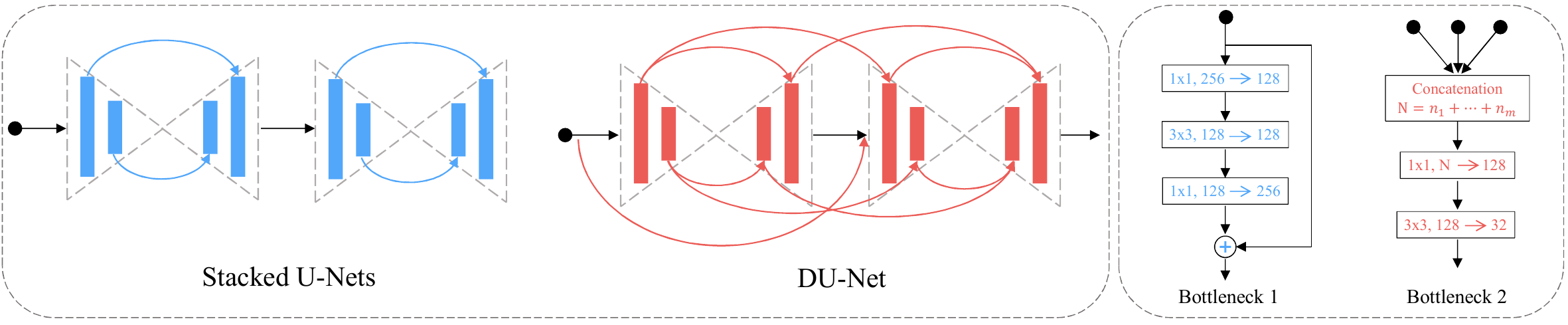}
\caption{Illustration of stacked U-Nets and DU-Net. Stacked U-Nets has skip connections only within each U-Net. In contrast, DU-Net also connects blocks with the same semantic meanings across different U-Nets. The feature reuse could significantly reduce the size of bottleneck in each block, as shown in the right figure. Consequently, with the same number of U-Nets, DU-Net has only 30\% parameters of stacked U-Nets.}
\label{fig:framework}
\end{figure*}

We propose densely connected U-Nets (DU-Net) to address this issue. The key idea is to directly connect blocks of the same semantic meanings, {\it i.e.} having the same resolution in either top-down or bottom-up context, from any U-Net to all subsequent U-Nets. Please refer to Fig. \ref{fig:framework} for an illustration. The dense connectivity is similar to DenseNet \cite{huang2016densely} but generalizing the design philosophy from feature to semantic level. It encourages information flow as well as feature reuse ``globally'' across the stacked U-Nets, yielding improved localization accuracy. 

Yet there are critical issues in designing DU-Net: 1) The number of parameters would have a quadratic growth since $n$ stacked U-Nets could generate $O(n^2)$ connections. 2) A naive implementation may allocate new memory for every connection, making the training highly expensive and limiting the maximum depth of DU-Nets. 


Our solution to those efficiency issues is threefold. {\bf First}, instead of connecting all stacked U-Nets, we only connect a U-Net to its $K$ successors. We name it as the $order$-$K$ connectivity, which aims to balance the fitting accuracy and parameter efficiency by cutting off long-distance connections. {\bf Second}, we employ a memory-efficient implementation in training. The key idea is to reuse a pre-allocated memory so all connected blocks could share the same memory. Compared with the naive implementation, this strategy makes it possible to train a very deep DU-Net (actually, $2\times$ deeper). {\bf Third}, to further improve the efficiency, we investigate an iterative design that may reduce the model size to one half. More specifically, the output of the first pass of the DU-Net is used as the input of the second pass, where detection or regression loss is applied as supervision. 

Besides shrinking the number of network parameters, we also study to further quantize each parameter. This motivates from the ubiquitous mobile applications. Although current mobile devices could carry models of dozens of MBs, deploying such networks requires high-end GPUs. However, quantized models could be accelerated by some specifically designed low-cost hardwares. Beyond only deploying models on mobile devices \cite{li2017deeprebirth}, training deep neural networks on distributed mobile devices emerges recently \cite{mcmahan2016communication}. To this end, we also try to quantize not only the model parameters but also its inputs (intermediate features) and gradients in training. This is the first attempt to investigate training landmark localizers using quantized inputs and gradients.

In summary, our key contributions are:
\begin{itemize}
    \item To the best of our knowledge, we are the first to propose quantized densely connected U-Nets for visual landmark localization, which largely improves the information flow and feature reuse at the semantic level.
    \item We propose the $order$-$K$ connectivity to balance accuracy and efficiency. It decreases the growth of model size from quadratic to linear by removing trivial connections. Experiments show it could reduce $\sim$70\% parameters of state-of-the-art landmark localizers.
    \item Very deep U-Nets can be trained using a memory-efficient implementation, where pre-allocated memory is reused by all connected blocks.
    \item We further investigate an iterative refinement that may cut down half of the model size, by forwarding DU-Net twice using either detection or regression supervision.
    \item Different from previous efforts of quantizing only the model parameters, we are the first to quantize their inputs and gradients for better training efficiency on landmark localization tasks. By choosing appropriate quantization bit-widths for weights, inputs and gradients, quantized DU-Net achieves $\sim$75\% training memory saving with comparable performance. 
    \item Exhaustive experiments are performed to validate DU-Net in different aspects. In both human pose estimation and face alignment, DU-Net demonstrates comparable localization accuracy and use $\sim$2\% model size compared with state-of-the-art methods.
\end{itemize}

\section{Related Work}
In this section, we review the recent developments on designing convolutional network architectures, quantizing the neural networks, human pose estimation and facial landmark localization.

{\bf Network Architecture.}
The identity mappings make it possible to train very deep ResNet \cite{he2016deep}. The popular stacked U-Nets \cite{newell2016stacked} are designed based on the residual modules. More recently, the DenseNet \cite{huang2016densely} outperforms the ResNet in the image classification task, benefitting from its dense connections. We would like to use the dense connectivity into multiple U-Nets.


{\bf Network Quantization.}
Training deep neural networks usually consumes a large amount of computational resources, which makes it hard to deploy on mobile devices. Recently, network quantization approaches \cite{courbariaux2016binarized,li2016ternary,zhou2016dorefa,wu2018training,rastegari2016xnor} offer an efficient solution to reduce the size of network through cutting down high precision operations and operands. In the recent binarized convolutional landmark localizer (BCLL)~\cite{bulat2017binarized} architecture, XNOR-Net \cite{rastegari2016xnor} was utilized for network binarization. However, BCLL only quantizes weights for inference and bring in real-value scaling factors. Due to its high precision demand in training, it cannot save training memory and improve training efficiency. To this end, we explore to quantize our DU-Net in training and inference simultaneously.

{\bf Human Pose Estimation.}
Starting from the DeepPose \cite{toshev2014deeppose}, CNNs based approaches \cite{wei2016convolutional,carreira2016human,bulat2016human,pishchulin2016deepcut,insafutdinov2016deepercut,lifshitz2016human,belagiannis2017recurrent,zhao2018learning} become the mainstream in human pose estimation and prediction. Recently, the architecture of stacked hourglasses \cite{newell2016stacked} has obviously beaten all the previous ones in terms of usability and accuracy. Therefore, all recent state-of-the-art methods \cite{chu2017multi,yang2017learning,yu2017adversarial,peng2018jointly} build on its architecture. They replace the residual modules with more sophisticated ones, add graphical models to get better inference, or use an additional network to provide adversarial supervisions or do adversarial data augmentation \cite{peng2018jointly}. In contrast, we design a simple yet very effective connectivity pattern for stacked U-Nets.

{\bf Facial Landmark Localization.}
Similarly, CNNs have largely reshaped the field of facial landmark localization. Traditional methods could be easily outperformed by the CNNs based \cite{zhang2014coarse,zhang2014facial,lv2017deep,peng2016recurrent,peng2018red}. In the recent Menpo Facial Landmark Localization Challenge \cite{zafeiriou2017menpo}, stacked hourglasses \cite{newell2016stacked} achieves state-of-the-art performance. The proposed $order$-$K$ connected U-Nets could produce even better results but with much fewer parameters.

\section{Our Method}
In this section, we first introduce the DU-Net after recapping the stacked U-Nets \cite{newell2016stacked}. Then we present the $order$-$K$ connectivity to improve its parameter efficiency, an efficient implementation to reduce its training memory,
and an iterative refinement to make it more parameter efficient.
Finally, network quantization is utilized to further reduce training memory and model size.
\subsection{DU-Net}
A U-Net contains top-down, bottom-up blocks and skip connections between them. Suppose multiple U-Nets are stacked together, for the $\ell^{th}$ top-down and bottom-up blocks in the $n^{th}$ U-Net, we use $f_\ell^n(\cdot)$ and $g_\ell^n(\cdot)$ to denote their non-linear transformations. Their outputs are represented by ${\bf x}_\ell^n$ and ${\bf y}_\ell^n$. $f_\ell^n(\cdot)$ and $g_\ell^n(\cdot)$ comprise operations of Convolution (Conv), Batch Normalization (BN) \cite{ioffe2015batch}, rectified linear units (ReLU) \cite{glorot2011deep}, and pooling.

{\bf Stacked U-Nets.} The feature transitions at the $\ell^{th}$ top-down and bottom-up blocks of the $n^{th}$ U-Net are:

\begin{equation}\label{eq:hg}
    {\bf x}_\ell^n = f_\ell^n({\bf x}_{\ell-1}^n), {\bf y}_\ell^n = g_\ell^n({\bf y}_{\ell-1}^n+{\bf x}_\ell^n).
\end{equation}
The skip connections only exist locally within each U-Net, which may restrict that information flows across U-Nets.

{\bf DU-Net.} To make information flow efficiently across stacked U-Nets, we propose a global connectivity pattern. Blocks at the same locations of different U-Nets have direct connections. Hence, we refer to this densely connected U-Nets architecture as {\it DU-Net}. Figure \ref{fig:framework} gives an illustration. Mathematically, the feature transitions at the $\ell^{th}$ top-down and bottom-up blocks of the $n^{th}$ U-Net can be formulated as:
\begin{equation}\label{eq:dense-inputs}
    {\bf x}_\ell^n = f_\ell^n([{\bf x}_{\ell-1}^n, {\bf X}_\ell^{n-1}]), {\bf y}_\ell^n = g_\ell^n([{\bf y}_{\ell-1}^n,{\bf x}_\ell^n,{\bf Y}_\ell^{n-1}]),
\end{equation}
where ${\bf X}_\ell^{n-1}={\bf x}_\ell^0,{\bf x}_\ell^1,\cdots,{\bf x}_\ell^{n-1}$ are the outputs of the $\ell^{th}$ top-down blocks in all preceding U-Nets. Similarly, ${\bf Y}_\ell^{n-1}={\bf y}_\ell^0,{\bf y}_\ell^1,\cdots,{\bf y}_\ell^{n-1}$ represent the outputs from the $\ell^{th}$ bottom-up blocks. $[\cdots]$ denotes the feature concatenation, which could make information flow more efficiently than the summation operation in Equation \ref{eq:hg}. 

According to Equation \ref{eq:dense-inputs}, a block receives features not only from connected blocks in the current U-Net but also the output features of the same semantic blocks from all its preceding U-Nets. Note that this semantic level dense connectivity is a generalization of the dense connectivity in DenseNet \cite{huang2016densely} that connects layers only within each block.

\begin{figure}[t]
\minipage[t]{0.49\textwidth}
\centering
  \includegraphics[width=0.9\linewidth]{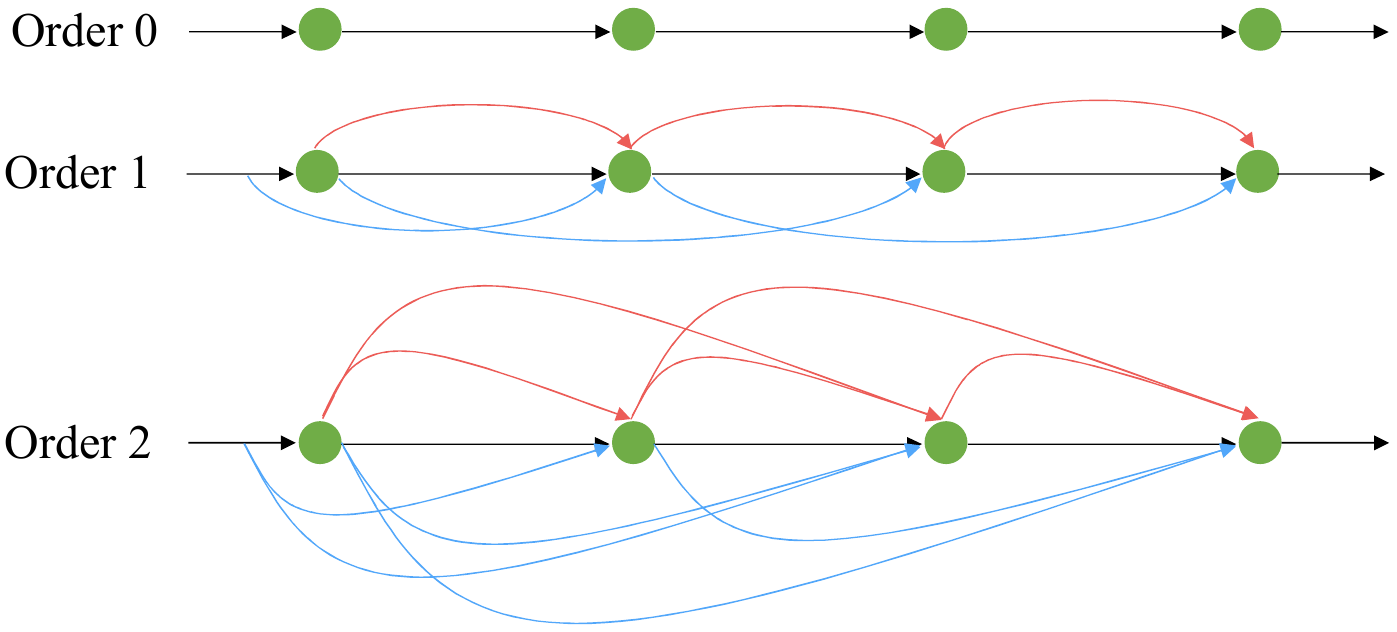}
  \caption{Illustration of $order$-$K$ connectivity. For simplicity, each dot represents one U-Net. The red and blue lines are the shortcut connections of inside semantic blocks and outside inputs. $Order$-$0$ connectivity ({\bf Top}) strings U-Nets together only by their inputs and outputs, i.e. stacked U-Nets. $Order$-$1$ connectivity ({\bf Middle}) has shortcut connections for adjacent U-Nets. Similarly, $order$-$2$ connectivity ({\bf Bottom}) has shortcut connections for 3 nearby U-Nets.}
\label{fig:$order$-$K$-illustr}
\endminipage \hfill
\minipage[t]{0.49\textwidth}
\centering
  \includegraphics[width=0.9\linewidth]{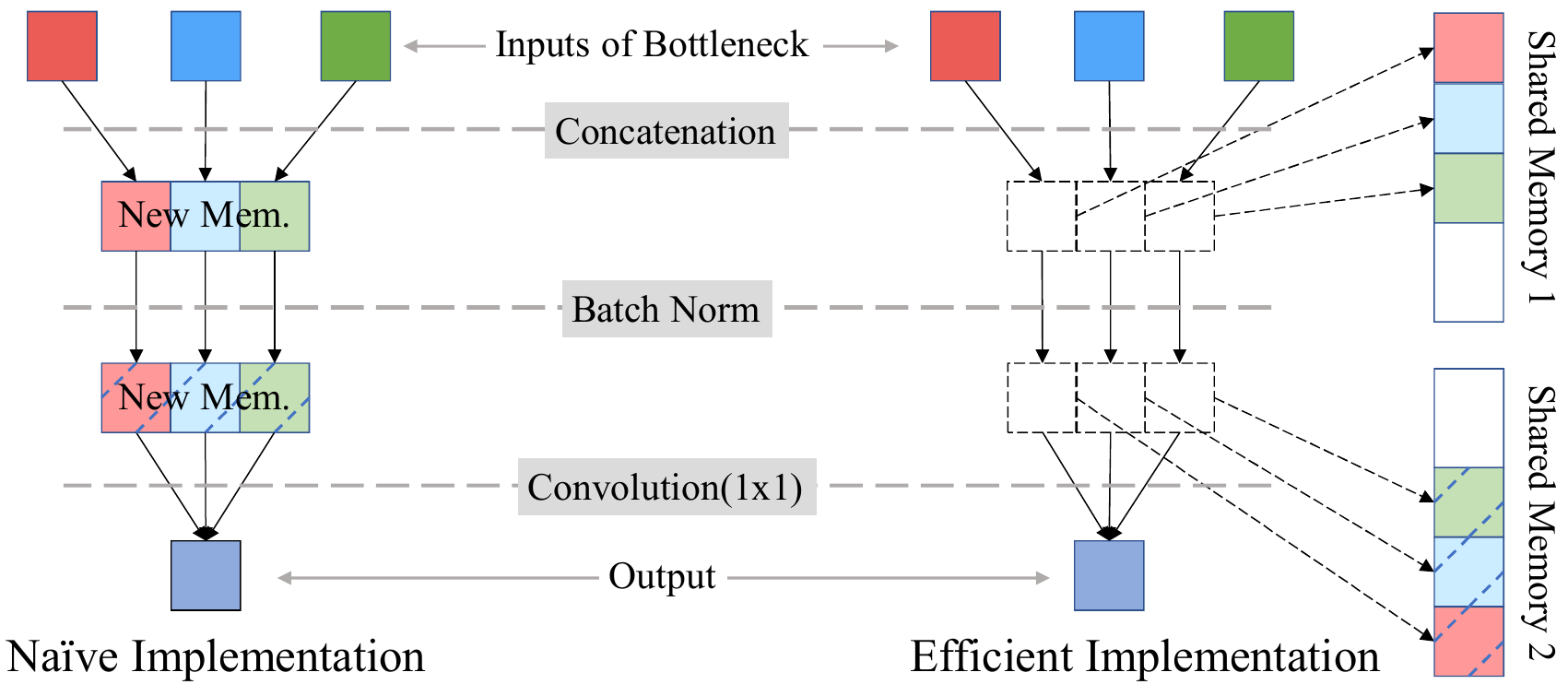}
  \caption{Illustration of memory efficient implementation. It is for the Concat-BN-ReLU-Conv($1\times 1$) in each bottleneck structure. ReLU is not shown since it is an inplace operation with no memory request. The efficient implementation  pre-allocates fixed memory space to store the concatenated and normalized features of connected blocks. In contrast, the naive implementation always allocates new memories for them, causing high memory consumption.}
  \label{fig:memory-efficient} \hfill
\endminipage
\end{figure}


\subsection{${\bf Order}$-${\bf K}$ Connectivity}
In the above formulation of DU-Net, we connect blocks with the same semantic meanings across all U-Nets. The connections would have quadratic growth depth-wise. To make DU-Net parameter efficient, we propose to cut off some trivial connections. For compensation, we add an intermediate supervision at the end of each U-Net. The intermediate supervisions, as the skip connections, could also alleviate the gradient vanish problem. Mathematically, the features  ${\bf X}_\ell^{n-1}$ and ${\bf Y}_\ell^{n-1}$ in Equation \ref{eq:dense-inputs} turns into 
\begin{gather}
    {\bf X}_{\ell}^{n-1}={\bf x}_{\ell}^{n-k},\cdots,{\bf x}_{\ell}^{n-1},\\
    {\bf Y}_{\ell}^{n-1}={\bf y}_{\ell}^{n-k},\cdots,{\bf y}_{\ell}^{n-1},
\end{gather}
where $0\leq k\leq n$ represents how many preceding nearby U-Nets connect with the current one. $k=n$ or $k=0$ would result in the stacked U-Nets or fully densely connected U-Nets. A medium order could reduce the growth of DU-Net parameters from quadratic to linear. Therefore, it largely improves the parameter efficiency of DU-Net and could make DU-Net grow several times deeper.

The proposed $order$-$K$ connection has similar philosophy as the Variable Order Markov (VOM) models \cite{begleiter2004prediction}. Each U-Net can be viewed as a state in the Markov model. The current U-Net depends on a fixed number of preceding nearby U-Nets, instead of preceding either only one or all U-Nets. In this way, the long-range connections are cut off. Figure \ref{fig:$order$-$K$-illustr} illustrates connections of three different orders. In Figure \ref{fig:$order$-$K$-illustr}, the connections above the central axes follow VOM patterns of $order$-$0$, $order$-$1$ and $order$-$2$ whereas the central axes together with connections below them follow VOM patterns of $order$-$1$, $order$-$2$ and $order$-$3$.

Dense connectivity is a special case of $order$-$K$ connectivity on the limit of $K$. For small $K$, $order$-$K$ connectivity is much more parameter efficient. But fewer connections may affect the prediction accuracy of very deep DU-Net. To make DU-Net have both high parameter efficiency and prediction accuracy, we propose to use $order$-$K$ connectivity in conjunction with intermediate supervisions. In contrast, DenseNet \cite{huang2016densely} has only one supervision at the end. Thus, it cannot effectively take advantage of $order$-$K$ connectivity.


\subsection{Memory Efficient Implementation}
Benefitting from the $order$-$K$ connectivity, our DU-Net is quite parameter efficient. However, a naive implementation would prevent from training very deep DU-Net, since every connection would make a copy of input features. To reduce the training memory, we follow the efficient implementation \cite{pleiss2017memory}. More specifically, concatenation operations of the same semantic blocks in all U-Nets share a memory allocation and their subsequent batch norm operations share another memory allocation. Suppose a DU-Net includes $N$ U-Nets each of which has $L$ top-down blocks and $L$ bottom-up blocks. We need to pre-allocate two memory space for each of $2L$ semantic blocks. For the $\ell^{th}$ top-down blocks, the concatenated features $[{\bf x}_{\ell-1}^1, {\bf X}_\ell^0], \cdots, [{\bf x}_{\ell-1}^{N-1}, {\bf X}_\ell^{N-2}]$ share the same memory space. Similarly, the concatenated features $[{\bf y}_{\ell-1}^0,{\bf x}_\ell^0], [{\bf y}_{\ell-1}^1,{\bf x}_\ell^1,{\bf Y}_\ell^0], \cdots, [{\bf y}_{\ell-1}^{N-1},{\bf x}_\ell^{N-1},{\bf Y}_\ell^{N-2}]$ in the $\ell^{th}$ bottom-up blocks share the same memory space.

In one shared memory allocation, later produced features would overlay the former features. Thus, the concatenations and their subsequent batch norm operations require to be re-computed in backward phase. Figure \ref{fig:memory-efficient} illustrates naive and efficient implementations.


\subsection{Iterative Refinement}
In order to further improve the parameter efficiency of DU-Net, we consider an iterative refinement. It uses only half of a DU-Net but may achieve comparable performance. In the iterative refinement, a DU-Net has two forward passes. In the first pass, we concatenate the inputs of the first and last U-Nets and merge them in a small dense block. Then the refined input is fed forward in the DU-Net again. Better output is expected because of the refined input. 

In this iterative pipeline, the DU-Net has two groups of supervisions in the first and second iterations. Both the detection and regression supervisions \cite{bulat2016human} are already used in the landmark detection tasks. However, there is no investigation how they compare with each other. To this end, we could try different combinations of detection and regression supervisions for two iterations. Our comparison could give some guidance for future research. 

\subsection{Network Quantization}

We aim at cutting down high precision operations and parameters both in training and inference stages of DU-Net. The bit-width of weights can be reduced to one or two bits through sign function or symmetrical threshold, whereas the layerwise gradients and inputs are quantized with linear mapping. In previous XNOR-Net \cite{rastegari2016xnor}, a scaling factor was introduced to approximate the real-value weight. However, calculating these float factor costs additional computational resources. To further decrease memory usage and model size, we try to remove the scaling factor and follow WAGE \cite{wu2018training} to quantize dataflow during training. More specifically, weights are binarized to -1 and 1 by the following equation:
\begin{equation} \label{eq:sign}
q(x) = sign(clip(x,-1,1))
\end{equation} 
or ternarized to -1, 0 and -1 by the a positive threshold $\delta$ as \cite{li2016ternary} presented, where $\delta \approx \frac{0.7}{n}\sum_{i=1}^{n} \left | w_i \right |$ provided that $w_i$ is initialized by Gaussian distributions. The dataflows, i.e. gradients and inputs, are quantized to $k$-bit values by the following linear mapping function:
\begin{equation} \label{eq:sign}
q(x,k) = clip(\sigma(k)\cdot  round(x\sigma(k)) -1+\sigma(k),1-\sigma(k))
\end{equation} 
Here, the unit distance $\sigma$ is calculated by $\sigma(k) = \frac{1}{2^{k-1}}$. In the following experiments, we explore different combinations of bit-widths to balance performance and memory consumption.


\section{Experiments}

\begin{figure*}[t!]
\centering
  \includegraphics[width=0.9\linewidth]{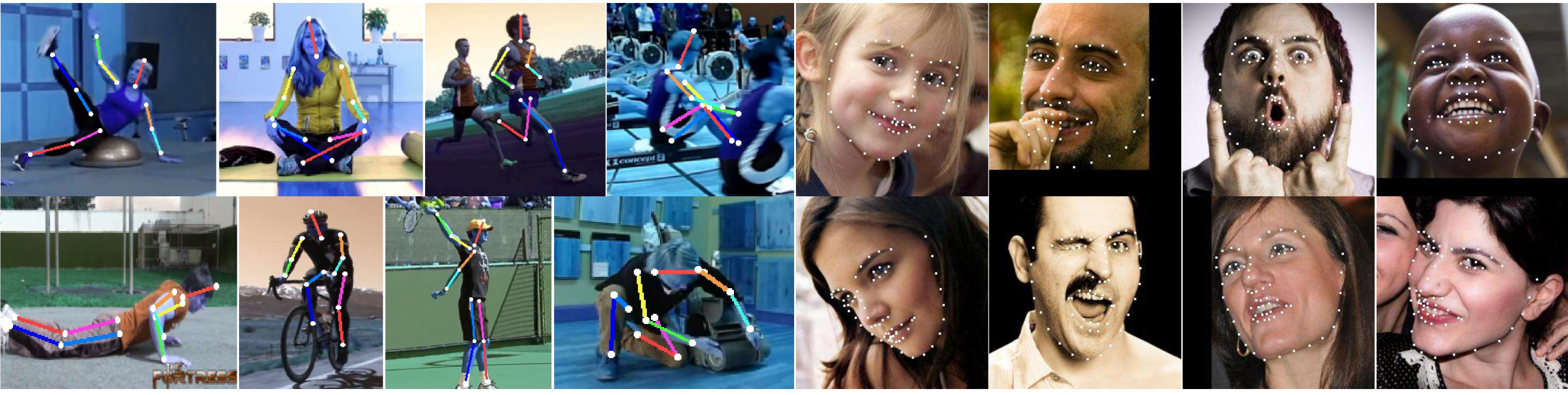}
\caption{Qualitative results of human pose estimation and facial landmark localization. DU-Net could handle a wide range of human poses, even with occlusions. It could also detect accurate facial landmarks with various head poses and expressions.}
\label{fig:pose-face-qualitive}
\end{figure*}

In this section, we first demonstrate the effectiveness of DU-Net through its comparison with the stacked U-Nets. Then we explore the relation between the prediction accuracy and $order$-$K$ connectivity. After that, we evaluate the iterative refinement to halve DU-Net parameters. Finally, we test the network quantization. Different combinations of bit-widths to find appropriate ones which balance accuracy, model size and memory consumption. The general comparisons are given at last. Some qualitative results are shown in Figure \ref{fig:pose-face-qualitive}.

{\bf Network.} The input resolution is normalized to 256$\times$256. Before the DU-Net, a Conv($7\times 7$) filter with stride 2 and a max pooling would produce 128 features with resolution 64$\times$64. Hence, the maximum resolution of DU-Net is 64$\times$64. Each block in DU-Net has a bottleneck structure as shown on the right side of Figure \ref{fig:framework}. At the beginning of each bottleneck, features from different connections are concatenated and stored in a shared memory. Then the concatenated features are compressed by the Conv($1\times 1$) to 128 features. At last, the Conv($3\times 3$) further produces 32 new features. The batch norm and ReLU are used before the convolutions. 

{\bf Training.} We implement the DU-Net using the PyTorch. The DU-Net is trained by the optimizer RMSprop. When training human pose estimators, the initial learning rate is $2.5\times 10^{-4}$ which is decayed to $5\times 10^{-5}$ after 100 epochs. The whole training takes 200 epochs. The facial landmark localizers are easier to train. Also starting from $2.5\times 10^{-4}$, its learning rate is divided by 5, 2 and 2 at epoch 30, 60 and 90 respectively. The above settings remain the same for quantized
DU-Net. In order to match the pace of dataflow, we set the same bit-width for gradients and inputs. We quantize dataflows and parameters all over the DU-Net except the first and last convolutional layers, since localization is a fine-grained task requires high precision of heatmaps.

{\bf Human Pose Datasets.} We use two benchmark human pose estimation datasets: MPII Human Pose \cite{andriluka14cvpr} and Leeds Sports Pose (LSP) \cite{johnson2010lsp}. The {\bf MPII} is collected from YouTube videos with a broad range of human activities. It has 25K images and 40K annotated persons, which are split into a training set of 29K and a test set of 11K. Following \cite{tompson2015efficient}, 3K samples are chosen from the training set as validation set. Each person has 16 labeled joints. The {\bf LSP} dataset contains images from many sport scenes. Its extended version has 11K training samples and 1K testing samples. Each person in LSP has 14 labeled joints. Since there are usually multiple people in one image, we crop around each person and resize it to 256x256. We also use scaling (0.75-1.25), rotation (-/+30) and random flip to augment the data.

{\bf Facial Landmark Datasets.} The experiments of the facial lanmark localization are conducted on the composite of HELEN, AFW, LFPW and IBUG which are re-annotated in the 300-W challenge \cite{sagonas2013300}. Each face has 68 landmarks. Following \cite{zhu2015face} and \cite{lv2017deep}, we use the training images of HELEN, LFPW and all images of AFW, totally 3148 images, as the training set. The testing is done on the common subset (testing images of HELEN and LFPW), challenge subset (all images from IBUG) and their union. We use the provided bounding boxes from the 300-W challenge to crop faces. The same augmentations of scaling and rotation as in human pose estimation are applied.

{\bf Metric.} We use the standard metrics in both human pose estimation and face alignment. Specifically, Percentage of Correct Keypoints (PCK) is used to evaluate approaches for human pose estimation. And the normalized mean error (NME) is employed to measure the performance of localizing facial landmarks. Following the convention of 300-W challenge, we use the inter-ocular distance to normalize mean error. For network quantization, we propose the balance index (BI) to examine the trade-off between performance and efficiency.


\begin{table}[t]
\minipage[t]{0.49\textwidth}
\centering
\caption{$Order$-1 DU-Net {\it v.s.} stacked U-Nets on MPII validation set measured by PCKh(\%) and parameter number. $Order$-1 DU-Net achieves comparable performance as stacked U-Nets. But it has only about 30\% parameters of stacked U-Nets. The feature reuse across U-Nets make each U-Net become light-weighted.}\label{tb:hg-vs-du-nets}
\begin{adjustbox}{width=1\textwidth}
\begin{tabular}{lccc}
\toprule
\multirow{2}{*}{Method} & PCKh & \#  & Parameter \\
& & Parameters & Ratio\\
\hline
Stacked U-Nets(16) & - & 50.5M & 100\% \\
DU-Net(16) & 89.9 & 15.9M & 31.5\% \\
\hline
Stacked U-Nets(8) & 89.3 & 25.5M & 100\%\\
DU-Net(8) & 89.5 & 7.9M & 31.0\% \\
\hline
Stacked U-Nets(4) & 88.3 & 12.9M & 100\%\\
DU-Net(4) & 88.2 & 3.9M & 30.2\% \\
\bottomrule
\end{tabular}
\end{adjustbox}
\endminipage \hfill
\minipage[t]{0.49\textwidth}
\centering
\caption{NME(\%) on 300-W using $Order$-1 DU-Net(4) with iterative refinement, detection and regression supervisions. The top two and bottom three rows are non-iterative and iterative results. Iterative refinement could lower localization errors. Besides, the regression supervision outperforms the detection supervision.}
\label{tb:iter}
\begin{adjustbox}{width=1\textwidth}
\begin{tabular}{lcccc}
\toprule
\multirow{2}{*}{Method} & Easy  & Hard  & Full & \#\\
&Subset & Subset & Set & Para.\\
\hline
Detection only & 3.63 & 5.60 & 4.01 & 3.9M\\
Regression only &  2.91 & 5.12 & 3.34 & 3.9M\\
\hline
Detection Detection & 3.52 & 5.59 & 3.93 & 4.1M\\
Detection Regression & 2.95 & 5.12 & 3.37 & 4.1M\\
Regression Regression  & 2.87 & 4.97 & 3.28 & 4.1M\\
\bottomrule
\end{tabular} \hfill
\end{adjustbox}
\endminipage
\end{table}


\subsection{DU-Net {\it vs.} Stacked U-Nets}
To demonstrate the advantages of DU-Net, we first compare it with traditional stacked U-Nets. This experiment is done on the MPII validation set. All DU-Nets use the $order$-$1$ connectivity and intermediate supervisions. Table \ref{tb:hg-vs-du-nets} shows three pairs of comparisons with 4, 8 and 16 U-Nets. Both their PCKh and number of convolution parameters are reported. We could observe that, with the same number of U-Nets, DU-Net could obtain comparable or even better accuracy. More importantly, the number of parameters in DU-Net is decreased by about 70\% of that in stacked U-Nets. The feature reuse across U-Nets make each U-Net in DU-Net become light-weighted. Besides, the high parameter efficiency makes it possible to train 16 $order$-$1$ connected U-Nets in a 12G GPU with batch size 16. In contrast, training 16 stacked U-Nets is infeasible. Thus, $order$-$1$ together with intermediate supervisions could make DU-Net obtain accurate prediction as well as high parameter efficiency, compared with stacked U-Nets.


\subsection{Evaluation of $Order$-$K$ connectivity}
\begin{figure}[t]
\minipage[t]{0.49\textwidth}
\centering
  \includegraphics[width=1\linewidth]{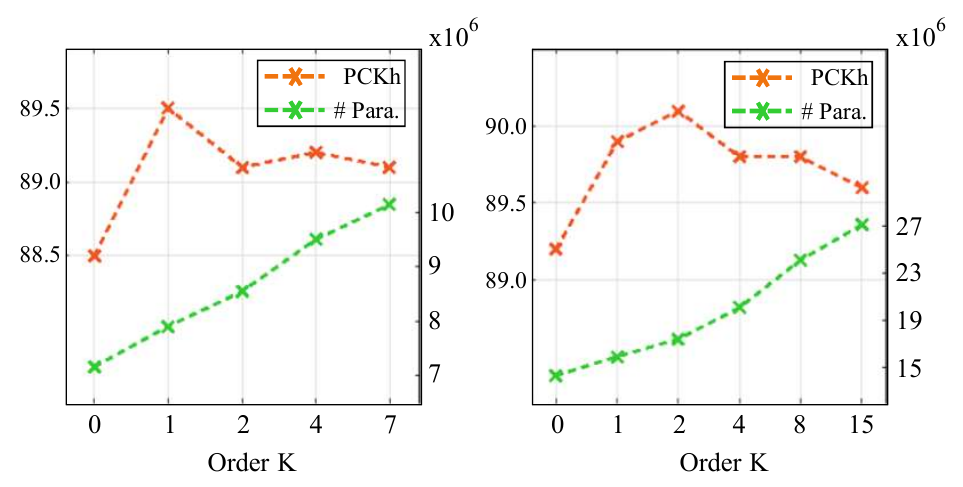}
  \caption{Relation of PCKh(\%), \# parameters and $order$-$K$ connectivity on MPII validation set. The parameter number of DU-Net grows approximately linearly with the order of connectivity. However, the PCKh first increases and then decreases. A small order 1 or 2 would be a good balance for prediction accuracy and parameter efficiency.}
\label{fig:exp-order-k}
\endminipage \hfill
\minipage[t]{0.49\textwidth}
\centering
  \includegraphics[width=0.9\linewidth]{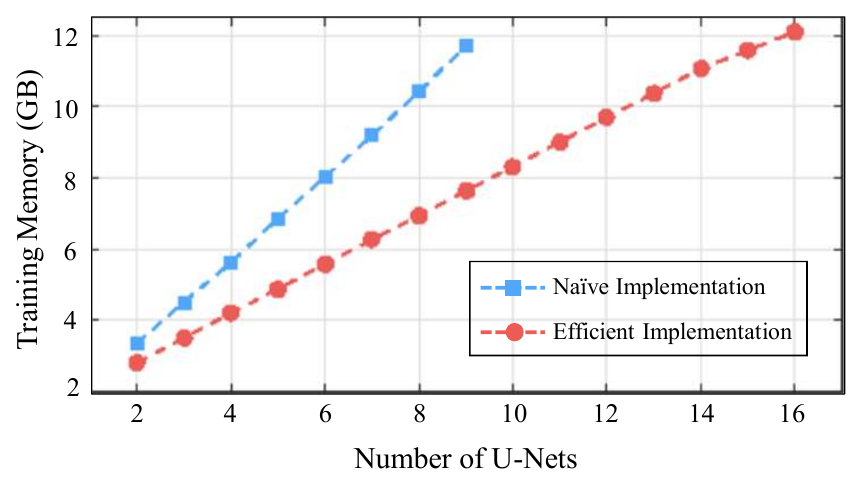}
  \caption{Naive implementation {\it v.s.} memory-efficient implementation. The $order$-$1$ connectivity, batch size 16 and a 12GB GPU are used. The naive implementation can only support 9 U-Nets at most. In contrast, the memory-efficient implementation allows to train 16 U-Nets, which nearly doubles the depth of DU-Net.}
  \label{fig:exp-naive-vs-efficient} \hfill
\endminipage
\end{figure}

The proposed $order$-$K$ connectivity is key to improve the parameter efficiency of DU-Net. In this experiment, we investigate how the PCKh and convolution parameter number change along with the order value. Figure \ref{fig:exp-order-k} gives the results from MPII validation set. The left and right figures show results of DU-Net with 8 and 16 U-Nets. It is clear that the convolution parameter number increases as the order becomes larger. However, the left and right PCKh curves have a similar shape of first increasing and then decreasing. $Order$-$1$ connectivity is always better than $order$-$0$. 

However, very dense connections may not be a good choice, which is kind of counter-intuitive. This is because the intermediate supervisions already provide additional gradients. Too dense connections make gradients accumulate too much, causing the overfitting of training set. Further evidence of overfitting is shown in Table \ref{tb:overfitting}. The $order$-7 connectivity has the higher training PCKh the $order$-1 in all training epochs. But its validation PCKh is a little lower in the last training epochs. Thus, small orders are recommended in DU-Net.



\begin{table}[t]
\minipage[t]{0.49\textwidth}
\centering
\setlength\tabcolsep{4pt}
\caption{$Order$-1 DU-Net(8) {\it v.s.} $order$-7 DU-Net(8), measured by training and validation PCKhs(\%) on MPII. $Order$-7 DU-Net(8) overfits the training set a little bit. Its validation PCKh is lower at last, though it always has higher training PCKh.}\label{tb:overfitting}
\begin{adjustbox}{width=1\textwidth}
\begin{tabular}{l|cccc}
\toprule
\multicolumn{5}{c}{PCKh on training set}\\
\hline
Epoch & 1 & 50 & 100 & 150 \\
\hline
$Order$-1 DU-Net(8) & 20.3 & 83.2 & 87.7 & 91.7 \\
$Order$-7 DU-Net(8) & {\bf 25.2} & {\bf 84.7} & {\bf 89.3} & {\bf 93.1} \\
\hline
\multicolumn{5}{c}{PCKh on validation set}\\
\hline
Epoch & 1 & 50 & 100 & 150 \\
\hline
$Order$-1 DU-Net(8) & 29.4 & 82.8 & {\bf 85.7} & {\bf 87.1}\\
$Order$-7 DU-Net(8) & {\bf 36.6} & {\bf 84.0} & 85.1 & 86.7\\
\bottomrule
\end{tabular}
\end{adjustbox}
\endminipage \hfill
\minipage[t]{0.49\textwidth}
\centering
\caption{Iterative $order$-1 DU-Net(4) {\it v.s.} non-iterative $order$-1 DU-Net(8) on 300-W measured by NME(\%). Iterative DU-Net(4), with few additional parameters on DU-Net(4), achieves comparable performance as DU-Net(8). Thus, the iterative refinement has the potential to halve parameters of DU-Net but still maintain comparable performance.}
\begin{adjustbox}{width=1\textwidth}
\label{tb:iter4-vs-8}
\begin{tabular}{lcccc}
\toprule
\multirow{2}{*}{Method} & Easy  & Hard  & Full & \#\\
&Subset & Subset & Set & Parameters\\
\hline
DU-Net(4) &  2.91 & 5.12 & 3.34 & 3.9M\\
Iter. DU-Net(4)  & 2.87 & 4.97 & 3.28 & 4.1M\\
DU-Net(8) & 2.82 & 5.07 & 3.26 & 7.9M\\
\bottomrule
\end{tabular} \hfill
\end{adjustbox}
\endminipage
\end{table}

\subsection{Evaluation of Efficient Implementation}

The memory-efficient implementation makes it possible to train very deep DU-Net. Figure \ref{fig:exp-naive-vs-efficient} shows the training memory consumption of both naive and memory-efficient implementations of DU-Net with order-1 connectivity. The linear growths of training memory along with number of U-Nets is because of the fixed order connectivity. But the memory growth of efficient implementation is much slower than that of the naive one. With batch size 16, we could train a DU-Net with 16 U-Nets in 12GB GPU. Under the same setting, the naive implementation could accept only 9 U-Nets.

\subsection{Evaluation of Iterative Refinement}
The iterative refinement is designed to make DU-Net more parameter efficient. First, experiments are done on the 300-W dataset using DU-Net(4). Results are shown in Table \ref{tb:iter}. For both detection and regression supervisions, adding an iteration could lower the localization errors, demonstrating effectiveness of the iterative refinement. Meanwhile, the model parameters only increase 0.2M, making DU-Net even more parameter efficient. Besides, the regression supervision outperforms the detection one no matter in the iterative or non-iterative setting, making it a better choice for landmark localization. 

Further, we compare iterative DU-Net(4) with non-iterative DU-Net(8). Table \ref{tb:iter4-vs-8} gives the comparison. We could find that, the iterative DU-Net(4) could obtain comparable NME as DU-Net(8). However, DU-Net(8) has double parameters of  DU-Net(4) whereas iterative DU-Net(4) increases only 0.2M additional parameters on DU-Net(4).

\subsection{Evaluation of Network Quantization}

Through network quantization, high precision operations and parameters can be efficiently represented by a few discrete values. 
In order to find appropriate choices of bit-widths, we try a series of bit-width combinations on the 300-W dataset based on $order$-$1$ DU-Net(4). The performance and balance ability of these combinations on several methods are shown in Table \ref{tb:IWG-QUAN}, where DU-Net(4) is DU-Net with 4 blocks, BW and TW respectively represents binarized weight and ternarized weight without $\alpha$, BW-$\alpha$ is binarized weight with float scaling factor $\alpha$, the suffix QIG means quantized inputs and gradients. 

For mobile devices with limited computational resources, slightly performance drop is tolerable provided that corresponding large efficiency enhancement. For the evaluation purpose, we propose a balance index (BI) to better examine the trade-off between performance and efficiency: 
\begin{equation} \label{eq:bi}
BI = NME^2 \cdot TM \cdot MS  
\end{equation} 
where $TM$ and $MS$ is respectively short for training memory and model size compression ratios to the original network without quantization. The square of $NME$ is calculated in the above formula to emphasize the prior importance of performance. For BI, the smaller the value, the better the ability of balance.

According to Table \ref{tb:IWG-QUAN}, BW-QIG(818) could achieve the best balance between performance and model efficiency among all the combinations. BW-QIG(818) could reduce more than 4$\times$ training memory and 32$\times$ model size while reach a better performance than TSR \cite{lv2017deep}. Besides, BW-$\alpha$-QIG(818),  BW-QIG(616) and TW-QIG(626) also have small balance index. Among all the combinations, the binarized network with scaling factor $\alpha$, i.e. BW-$\alpha$ gets the closest error to the original network DU-Net(4).

For BW-$\alpha$-QIG(818), the performance is not better than BW-QIG(818). This is mainly because that BW-$\alpha$ is heavily rely on the parameter $\alpha$. However, the quantization of dataflow could reduce the approximation ability of $\alpha$. TW and TW-QIG usually gets better results than BW and BW-QIG, since they have more choices in terms of weight value. The above results proves the effectiveness of network quantization, yet a correct combination of bit-widths is a crucial factor.

\begin{table}[t!]
\begin{center}
\caption{Performance and balance ability of different combinations of bit-width values on the 300-W dataset measured by NME(\%), all quantized networks are based on $order$-$1$ DU-Net(4). BW and TW is short for binarized and ternarized weight, $\alpha$ represents float scaling factor, QIG is short for quantized inputs and gradients. $Bit_I$, $Bit_W$, $Bit_G$ represents the bit-width of inputs, weights, gradients respectively. Training memory and model size is represented by the compression ratio to the original DU-Net(4). Balance index is calculated by equation \ref{eq:bi}. Comparable error rate could be achieved by binarized the model parameters. Further quantizing the inputs and gradients could substantially reduce the training memory with some increase of detection error. The balance index is a indicator for balancing the quantization and accuracy.}

\small



\label{tb:IWG-QUAN}
\begin{tabular}{lccccccccc}
\toprule
\multirow{2}{*}{Method} & {$Bit_I$}  & $Bit_W$  & $Bit_G$ & NME(\%) & NME(\%) & NME(\%) & Training  &  Model & Balance\\
& & & & Full set & Easy set & Hard set & Memory & Size & Index\\

\hline
DU-Net(4)  &  32  & 32  & 32  & 3.38 & 2.95 & 5.13 & 1.00 & 1.00 & 11.4 \\
\hline
BW-QIG     &  6   & 1   & 6   & 5.93 & 5.10 & 9.34 & 0.17 & 0.03 & 0.18	\\
BW-QIG     &  8   & 1   & 8   & 4.30 & 3.67 & 6.86 & 0.25 & 0.03 & \bf{0.14}	    \\
BW-$\alpha$-QIG    &  8   & 1   & 8   & 4.47 & 3.75 & 7.40 & 0.25 & 0.03 & 0.15  \\
BW         &  32  & 1   & 32  & 3.75 & 3.20 & 5.99 & 1.00 & 0.03 & 0.42	\\
BW-$\alpha$        &  32  & 1   & 32  & {\bf 3.58} & 3.12 & 5.45 & 1.00 & 0.03 & 0.38  \\
TW         &  32  & 2   & 32  & 3.73 & 3.21 & 5.85 & 1.00 & 0.06 & 0.83  \\
TW-QIG     &  6   & 2   & 6   & 4.27 & 3.70 & 6.59 & 0.17 & 0.06 & 0.19	\\
TW-QIG     &  8   & 2   & 8   & 4.13 & 3.55 & 6.50 & 0.25 & 0.06 & 0.26   \\
\bottomrule
\end{tabular}

\end{center}
\end{table}

\begin{table}[htb]
\begin{center}
\small
\caption{Comparison of convolution parameter number (Million) and model size (Megabyte) with state-of-the-art methods. DU-Net(16) has 27\%-62\% parameters of other methods. Its binarized version DU-Net-BW-$\alpha$(16) has less than {\bf 2\%} model size.}\label{tb:para-num}
\setlength\tabcolsep{0.5pt}
\begin{tabular}{lccccc|cc}
\toprule
\multirow{2}{*}{Method} & Yang  & Wei & Bulat  & Chu & Newell  & $Order$-1 & $Order$-1 DU-\\
& {\it et al.}\cite{yang2017learning} & {\it et al.}\cite{wei2016convolutional}  
& {\it et al.}\cite{bulat2016human} & {\it et al.}\cite{chu2017multi} & {\it et al.}\cite{newell2016stacked} & DU-Net(16) & Net-BW-$\alpha$(16)\\
\hline
\# Parameters & 28.0M & 29.7M & 58.1M & 58.1M & 25.5M & {\bf 15.9M} & {\bf 15.9M}\\
Model Size & 110.2MB & 116.9MB & 228.7MB & 228.7MB & 100.5MB & 62.6MB & {\bf 2.0MB}\\

\bottomrule
\end{tabular}
\end{center}
\end{table}

\subsection{Comparison with State-of-the-art Methods}


{\bf Human Pose Estimation.}
Tables \ref{tb:mpii} and \ref{tb:lsp} show comparisons of human pose estimation on MPII and LSP test sets. The $order$-$1$  DU-Net-BW-$\alpha$(16) achieves comparable state-of-the-art performances. In contrast, as shown in Table \ref{tb:para-num}, it has only 27\%-62\% parameters and less than 2\% model size of other recent state-of-the-art methods. The DU-Net is concise and simple. Other state-of-the-art methods use stacked U-Nets with either sophisticated modules \cite{yang2017learning}, graphical models \cite{chu2017multi} or adversarial networks \cite{yu2017adversarial}.

{\bf Facial Landmark Localization.}
The DU-Net is also compared with other state-of-the-art facial landmark localization methods on 300-W. Please refer to Table \ref{tb:300w}. We uses a smaller network $order$-1 DU-Net(8) than that in human pose estimation, since localizing the facial landmarks is easier. The $order$-1 DU-Net-BW-$\alpha$(8) gets comparable errors state-of-the-art method \cite{newell2016stacked}. However, $order$-1 DU-Net-BW-$\alpha$(8) has only $\sim$2\% model size.

\begin{table}[t!]
\begin{center}
\small
\setlength\tabcolsep{1.5pt}
\caption{PCKh(\%) comparison on MPII test sets. $Order$-$1$ DU-Net could achieve comparable performance as state-of-the-art methods. More importantly, DU-Net-BW-$\alpha$(16) has at least $\sim${\bf 30\%} parameters and at most $\sim${\bf 2\%} model size.}\label{tb:mpii}
\begin{tabular}{@{}lcccccccc@{}}
\toprule
Method & Head & Sho. & Elb. & Wri. & Hip & Knee & Ank. & Mean\\
\hline
Pishchulin \textit{et al.} ICCV'13 \cite{pishchulin2013strong} & 74.3 & 49.0 & 40.8 & 34.1 & 36.5 & 34.4 & 35.2 & 44.1\\
Tompson \textit{et al. } NIPS'14 \cite{tompson2014joint} & 95.8 & 90.3 & 80.5 & 74.3 & 77.6 & 69.7 & 62.8 & 79.6\\
Carreira \textit{et al.} CVPR'16 \cite{carreira2016human} & 95.7 & 91.7 & 81.7 & 72.4 & 82.8 & 73.2 & 66.4 & 81.3\\
Tompson \textit{et al.} CVPR'15 \cite{tompson2015efficient}& 96.1 & 91.9 & 83.9 & 77.8 & 80.9 & 72.3 & 64.8 & 82.0\\
Hu \textit{et al.} CVPR'16 \cite{hu2016bottom}& 95.0 & 91.6 & 83.0 & 76.6 & 81.9 & 74.5 & 69.5 & 82.4\\
Pishchulin \textit{et al.} CVPR'16 \cite{pishchulin2016deepcut}&94.1 & 90.2 & 83.4 & 77.3 & 82.6 & 75.7 & 68.6 & 82.4\\
Lifshitz \textit{et al.} ECCV'16 \cite{lifshitz2016human} & 97.8 & 93.3 & 85.7 & 80.4 & 85.3 & 76.6 & 70.2 & 85.0\\
Gkioxary \textit{et al.} ECCV'16 \cite{gkioxari2016chained} & 96.2 & 93.1 & 86.7 & 82.1 & 85.2 & 81.4 & 74.1 & 86.1\\
Rafi \textit{et al.} BMVC'16 \cite{rafi2016efficient} & 97.2 & 93.9 & 86.4 & 81.3 & 86.8 & 80.6 & 73.4 & 86.3\\
Belagiannis \textit{et al.} FG'17 \cite{belagiannis2017recurrent}&97.7 & 95.0 & 88.2 & 83.0 & 87.9 & 82.6 & 78.4 & 88.1\\
Insafutdinov \textit{et al.} ECCV'16 \cite{insafutdinov2016deepercut}&96.8 & 95.2 & 89.3 & 84.4 & 88.4 & 83.4 & 78.0 & 88.5\\
Wei \textit{et al.} CVPR'16 \cite{wei2016convolutional} & 97.8 & 95.0 & 88.7 & 84.0 & 88.4 & 82.8 & 79.4 & 88.5\\
Bulat \textit{et al.} ECCV'16 \cite{bulat2016human} & 97.9 & 95.1 & 89.9 & 85.3 & 89.4 & 85.7 & 81.7 & 89.7\\
Newell {\it et al.}  ECCV'16 \cite{newell2016stacked} & 98.2 & 96.3 & 91.2 & 87.1 & 90.1 & 87.4 & 83.6 & 90.9\\
Chu \textit{et al.} CVPR'17 \cite{chu2017multi} & {\bf 98.5} & 96.3 & 91.9 & {\bf 88.1} & {\bf 90.6} & {\bf 88.0} & {\bf 85.0} & {\bf 91.5}\\
\hline
$Order$-$1$ DU-Net(16) & 97.4  & {\bf 96.4}  & {\bf 92.1}  & 87.7  & 90.2  & 87.7 & 84.3 & 91.2\\
$Order$-$1$ DU-Net-BW-$\alpha$(16) & 97.6  & 96.4  & 91.7  & 87.3  & 90.4  & 87.3 & 83.8 & 91.0\\
\bottomrule
\end{tabular}
\end{center}
\end{table}

\begin{table}[t]
\begin{center}
\caption{NME(\%) comparison with state-of-the-art facial landmark localization methods on 300-W dataset. DU-Net-BW-$\alpha$ refers to the DU-Net with binarized weights and scaling factor $\alpha$ The binarized DU-Net obtains comparable performance with state-of-the-art method \cite{newell2016stacked}. But it has $\sim${\bf 50}$\times$ smaller model size.}\label{tb:300w}
\small
\setlength\tabcolsep{0.5pt}
\begin{tabular}{lccccccc|ccc}
\toprule
\multirow{2}{*}{Method} & CFAN  & Deep  & CFSS  
& TCDCN  & MDM  & TSR & HGs(4) & $Order$-1 & $Order$-1 DU-\\
& \cite{zhang2014coarse} & Reg \cite{shi2014deep} & \cite{zhu2015face} & \cite{zhang2014facial} &  \cite{trigeorgis2016mnemonic} & \cite{lv2017deep} & \cite{newell2016stacked} & DU-Net(8)& Net(8)-BW-$\alpha$\\
\hline
Easy subset  & 5.50 & 4.51  &  4.73 & 4.80 & 4.83  & 4.36 & 2.90 & {\bf 2.82} & 3.00\\ 
Hard subset  & 16.78 &  13.80  & 9.98 & 8.60 & 10.14 &  7.56 & 5.15 &{\bf 5.07} & 5.36\\
Full set   & 7.69 & 6.31 & 5.76 & 5.54 & 5.88 & 4.99 & 3.35 & {\bf 3.26} & 3.46\\
\bottomrule
\end{tabular}
\end{center}
\end{table}

\begin{table}[htb]
\begin{center}
\small
\caption{PCK(\%) comparison on LSP test set. The $Order$-$1$ DU-Net could also obtain comparable state-of-the-art performance. But DU-Net-BW-$\alpha$(16) has at most $\sim${\bf 70}\% fewer parameters and $\sim${\bf 50}$\times$ smaller model size than other state-of-the-art methods.}\label{tb:lsp}
\setlength\tabcolsep{1.5pt}
\begin{tabular}{@{}lcccccccc@{}}
\toprule
Method & Head & Sho. & Elb. & Wri. & Hip & Knee & Ank. & Mean\\
\hline
Belagiannis \textit{et al.} FG'17 \cite{belagiannis2017recurrent} & 95.2 & 89.0 & 81.5 & 77.0 & 83.7 & 87.0 & 82.8 & 85.2\\
Lifshitz \textit{et al.} ECCV'16 \cite{lifshitz2016human} & 96.8 & 89.0 & 82.7 & 79.1 & 90.9 & 86.0 & 82.5 & 86.7\\
Pishchulin \textit{et al.} CVPR'16 \cite{pishchulin2016deepcut} &  97.0 & 91.0 & 83.8 & 78.1 & 91.0 & 86.7 & 82.0 & 87.1\\
Insafutdinov \textit{et al.} ECCV'16 \cite{insafutdinov2016deepercut}& 97.4 & 92.7 & 87.5 & 84.4 & 91.5 & 89.9 & 87.2 & 90.1\\
Wei \textit{et al.} CVPR'16 \cite{wei2016convolutional}& 97.8 & 92.5 & 87.0 & 83.9 & 91.5 & 90.8 & 89.9 & 90.5\\
Bulat \textit{et al.} ECCV'16 \cite{bulat2016human}& 97.2 & 92.1 & 88.1 & 85.2 & 92.2 & 91.4 & 88.7 & 90.7\\
Chu \textit{et al.} CVPR'17 \cite{chu2017multi}& 98.1 & 93.7 & 89.3 & 86.9 &  93.4 & 94.0 & 92.5 & 92.6\\
Newell {\it et al.} ECCV'16 \cite{newell2016stacked} & {\bf 98.2} & 94.0 & 91.2 & 87.2 & 93.5 & {\bf 94.5} & 92.6 & 93.0\\
Yang \textit{et al.} ICCV'17 \cite{yang2017learning} & {\bf 98.3} & 94.5 & 92.2 & 88.9 & {\bf 94.4} & 95.0 & 93.7 & 93.9\\
\hline
$Order$-$1$ DU-Net(16) &  97.5 & {\bf 95.0} & {\bf 92.5} & {\bf 90.1} &  93.7 &  {\bf 95.2} & 94.2 & {\bf 94.0}\\
$Order$-$1$ DU-Net-BW-$\alpha$(16) &  97.8 & 94.3 & 91.8 & 89.3 &  93.1 &  94.9 & {\bf 94.4} & 93.6\\

\bottomrule
\end{tabular}
\end{center}
\end{table}

\section{Conclusion}
We have generalized the dense connectivity into the stacked U-Nets, resulting in a novel, simple and effective DU-Net. It connects blocks with the same semantic meanings in different U-Nets. $Order$-$K$ connectivity is proposed to improve its parameter efficiency. An iterative refinement is also introduced make it more parameter efficient. It could halve a DU-Net but achieves comparable accuracy.
Through network quantization, the training memory consumption and model size can further be reduced simultaneously.
 Experiments show the DU-Net could achieve state-of-the-art performances as other landmark localizers but with only  $\sim$30\% parameters, $\sim$2\% model size and $\sim$25\% training memory.
\section{Acknowledgment}
This work is partly supported by the Air Force Office of Scientific Research (AFOSR) under the Dynamic Data-Driven Application Systems Program, NSF 1763523, 1747778, 1733843 and 1703883 Awards.

\clearpage

\bibliographystyle{splncs04}
\bibliography{egbib}
\end{document}